\pdfoutput=1

\documentclass[11pt]{article}

\usepackage[final]{acl}

\usepackage{times}
\usepackage{latexsym}

\usepackage[T1]{fontenc}

\usepackage[utf8]{inputenc}

\usepackage{microtype}

\usepackage{inconsolata}

\usepackage{graphicx}

\usepackage{times}
\usepackage{latexsym}
\usepackage{longtable}
\usepackage{setspace}
\usepackage{lipsum}
\usepackage{etoolbox}
\usepackage{multicol}
\usepackage{listings}
\usepackage{fancyvrb}
\usepackage{color}
\usepackage{pifont}
\usepackage[utf8]{inputenc} 
\usepackage[T1]{fontenc}    
\usepackage[T1]{fontenc}    
\usepackage{url}            
\usepackage{booktabs}       
\usepackage{amsfonts}       
\usepackage{amsmath}
\usepackage{nicefrac}       
\usepackage{microtype}      
\usepackage{multirow}
\usepackage{caption}
\usepackage{subcaption}
\usepackage{xcolor}
\usepackage{float}
\usepackage{xspace}

\newcommand{\DOT}{D\texttt{0}T\xspace}

\title{Transforming Slot Schema Induction with\\ Generative Dialogue State Inference}

\author{
  James D. Finch
  \and
  Boxin Zhao
  \and
  Jinho D. Choi
  \\
  Department of Computer Science
  \\
  Emory University
  \\
  Atlanta, GA, USA
  \\
  \texttt{\{jdfinch, zinc.zhao, jinho.choi\}@emory.edu}
  \\
}

\begin{document}
\maketitle
\begin{abstract}
The challenge of defining a slot schema to represent the state of a task-oriented dialogue system is addressed by Slot Schema Induction (SSI), which aims to automatically induce slots from unlabeled dialogue data. Whereas previous approaches induce slots by clustering value spans extracted directly from the dialogue text, we demonstrate the power of discovering slots using a generative approach. By training a model to generate slot names and values that summarize key dialogue information with no prior task knowledge, our SSI method discovers high-quality candidate information for representing dialogue state. These discovered slot-value candidates can be easily clustered into unified slot schemas that align well with human-authored schemas. Experimental comparisons on the MultiWOZ and SGD datasets demonstrate that Generative Dialogue State Inference (\texttt{GenDSI}) outperforms the previous state-of-the-art on multiple aspects of the SSI task.
\end{abstract}

\begin{figure*}[htbp]
    \centering
    \includegraphics[width=0.9\textwidth]{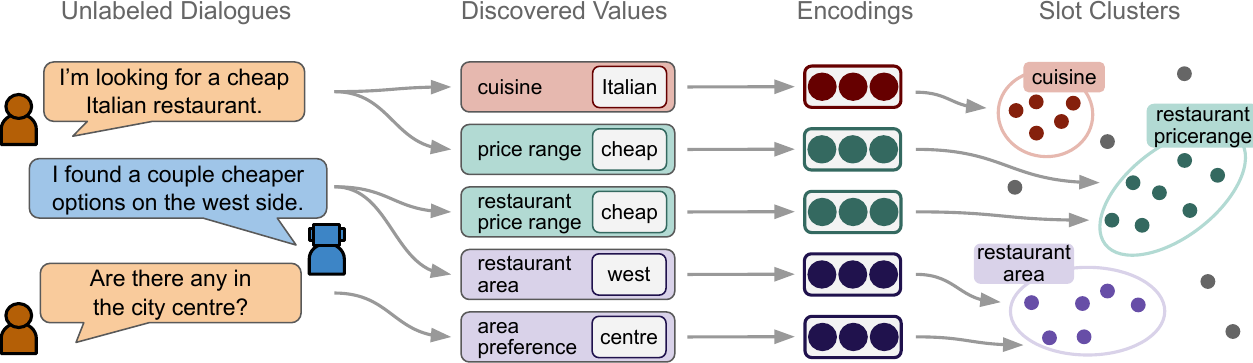}
    \caption{Overview of the \texttt{GenDSI} approach.}
    \label{fig:dsi-approach}
\end{figure*}

\section{Introduction}
\label{sec:ssi_intro}

Developing Task-Oriented Dialogue (TOD) systems presents the significant challenge of creating and maintaining a \textit{slot schema}, where each slot defines a type of information that is critical for successfully completing the dialogue task \cite{budzianowski_multiwoz_2018}. Traditionally slot schemas are handcrafted, but manually defining each slot is time-consuming, especially when task domains are complicated or the functionality of the dialogue system is frequently updated. To address this, Slot Schema Induction (SSI) has been proposed to automatically generate slot schemas from unlabeled dialogue data \cite{chen_unsupervised_2013, min_dialogue_2020}. This task facilitates the automatic analysis of dialogue structure \cite{qiu_structure_2022} and identifies key types of information that should be included in dialogue state representations \cite{min_dialogue_2020}. By reducing the need for manual schema creation, SSI expedites developing TOD systems for new application domains, and enables continual discovery of new slot types to improve the coverage of existing slot schemata.

The core challenge of SSI is identifying which information presented in unlabeled dialogue data is important for the task domain and should be included in the dialogue state. Once the important information values are identified, a second challenge is defining a minimal set of slots that captures the different types of information the values represent.
All previous work on SSI tackles these challenges in an explicit two-step process involving (1) candidate value identification and (2) inducing a slot schema by clustering candidate values into a set of slot clusters.
Identifying value candidates has been explored using tagging models trained on other tasks such as NER or SRL \cite{min_dialogue_2020, hudecek_discovering_2021, qiu_structure_2022, wu_semi-supervised_2022}, or using token attention distributions produced by a PLM to extract syntactic constituents \cite{yu_unsupervised_2022}. Inducing slots from value candidates has been explored using out-of-the-box clustering algorithms \cite{qiu_structure_2022}, multi-stage clustering pipelines specific to SSI \cite{hudecek_discovering_2021, wu_semi-supervised_2022, yu_unsupervised_2022}, or a neural latent variable model \cite{min_dialogue_2020}.

Unlike all previous approaches to SSI, we are the first to take a generative approach to value candidate identification.\footnote{The code, models, and data for our approach is publicly available at \url{https://github.com/emorynlp/GenDSI}.} Candidates are identified using a dialogue state generator model, which is trained to summarize the key task-related information in a given dialogue context as a set of state values. Crucially, this state generator also creates a slot name for each value, which serves as a candidate prediction of the name of the slot the value fills. Value candidates are then clustered in conjunction with these predicted slot names to induce a unified set of slots. 
The advantage of this approach is that the type semantics of each value candidate are concretely represented using slot name predictions, whereas previous approaches rely on the similarity of value encodings alone to cluster values into slot types. 
Predicting slot names has the additional benefit of enabling automatic naming of each slot cluster in the induced schema. 
We demonstrate the impact of these benefits by evaluating our approach on the MultiWOZ \cite{eric_multiwoz_2020} and SGD \cite{rastogi_towards_2020} datasets. Our SSI approach produces slot schemas that better match gold reference schemas when compared to the previous state-of-the-art (SoTA) approaches.

\section{Approach}
\label{sec:ssi_approach}

Our SSI approach, Generative Dialogue State Inference (\texttt{GenDSI}), induces a slot schema consisting of a set of slot clusters from an unlabeled set of dialogues. The induction procedure is performed in three stages (Fig.~\ref{fig:dsi-approach}). First, a dialogue state generator discovers value candidates for each turn in the dialogue data and jointly predicts a slot name with each value. Second, an encoding model produces a dense vector representation for each slot-value candidate. Finally, a clustering algorithm uses the encodings to filter and group candidates into a unified set of slot clusters. 

\paragraph{Dialogue State Generator} Our approach formulates the discovery of value candidates from unlabeled dialogue data as a sequence-to-sequence generation task. The input is a dialogue context $D_{*..t}$, and the output is a list of slot-value candidates $[(s_1, v_1),...,(s_k,v_k)]$ represented by the sequence format $s_1\text{:} v_1; s_2\text{:} v_2; ... s_k\text{:} v_k \text{[}eos\text{]}$. Each candidate includes a value $v_i$ that is inferred to belong to the dialogue state and a slot name prediction $s_i$ to represent the type of $v_i$. To enable the model to generate slot-value pairs that discover important dialogue state information without any prior knowledge of the task domain, we fine-tune a pretrained encoder-decoder transformer on TOD data that covers a large variety of domains. Section \ref{sec:dsg_evaluation} presents an evaluation of the dialogue state generator in which two different training datasets are compared.

\paragraph{Value Candidate Encoding} Each slot-value candidate $(s_i, v_i)$ produced by the dialogue state generator is encoded into a single dense vector representation $e_i$. To do this, we concatenate the slot name and value candidate with a separator to form a single token sequence $s_i\text{:}v_i$. We then use the SBERT encoder \cite{reimers_sentence-bert_2019} to independently encode each candidate sequence. 

\paragraph{Slot Clustering} Given a complete list of all slot-value candidates $[(s_1, v_1),...,(s_n,v_n)]$ produced by the dialogue state generator across all turns of the dialogue dataset, slot-value candidates are jointly filtered and grouped by applying the HDBSCAN algorithm \cite{mcinnes_hdbscan_2017} to the candidate encodings $[e_1, e_2, ..., e_n]$. As demonstrated in previous work \cite{yu_unsupervised_2022}, HDBSCAN is a suitable clustering algorithm because (1) it automatically discovers an appropriate number of slot clusters and (2) it filters out examples in low-density regions of the encoding space, which are likely to represent noisy candidates. The result is a set of slot clusters $[S_1, S_2, ..., S_k]$ where each cluster $S_i$ is a list of values that fill the slot it represents.

\section{State Generator Evaluation}
\label{sec:dsg_evaluation}

Since our SSI approach relies mainly on the dialogue state generator component to infer high-quality value candidates with appropriate slot names, we first conduct an evaluation of the performance of this component when discovering slot-values from dialogues in unseen task domains. 

\paragraph{Metrics} The dialogue state generator is evaluated by human judges, since discovered slot-value candidates are generated and many surface forms can be equally correct. We recruit three university students as volunteers to evaluate two key aspects of slot value candidate inferences. (1) \textit{Completeness} measures the proportion of dialogue turns for which all key information has been captured as slot-value candidates. 
(2) \textit{Correctness} measures the proportion of slot-value candidates that accurately represent specific information in their corresponding turns. 
Details of metrics are presented in Appx. \ref{appx:dsg_metrics}. This evaluation is performed using a custom annotation software, which was developed to optimize the efficiency of the annotation work. The interface is shown in Appx. \ref{appx:generator_evaluation_interface}. 

To validate our human evaluation metrics, inter-annotator agreement was calculated between the three human judges at 0.43 and 0.27 Krippendorff's Alpha for Correctness and Completeness respectively. Based on a manual review of the annotation disagreements for Completeness items, we believe the lower agreement occurs because judges are required to consider more information across an entire state update compared to judging the correctness of a single slot-value pair, leading to higher annotation difficulty and thus some noisy judgements.

\paragraph{Data} Since our goal is to train a dialogue state generator to discover slot-value candidates for unseen domains, we experiment with two domain-diverse datasets for training: SGD \cite{rastogi_towards_2020} and \DOT \cite{finch2024diverseeffectivesyntheticdata}. SGD is a popular TOD dataset that contains $20$ domains and $16,142$ dialogues, with gold dialogue state labels. \DOT is a recent dataset that was created using a fully automatic data generation method based on GPT-3.5 and GPT-4. It covers a large $1,003$ domains across $5,015$ dialogues, but it contains some noisy labels from automatic annotation. 

We adapt these datasets for slot-value candidate discovery by simply training on dialogue state \textit{updates} instead of full dialogue states, which represent only the slots that are filled by new values. This avoids training the dialogue state generator to predict empty slots, which are not useful for candidate discovery. Additionally, the special request value "?" is removed from \DOT state updates. We also replace each slot name in the SGD training split with a random synonymous name from SGD-X \cite{lee_sgd-x_2022}, as we found this augmentation to improve performance.

Both SGD and \DOT are also used as evaluation data by randomly sampling $60$ turns from their test splits, each from a unique dialogue. Crucially, we \textit{only} sample turns from domains \textit{not} included in the training split. Since the \DOT dataset has no native splits for training and testing, we randomly sample $100$ domains out of the total $1,003$ to be held-out for evaluation. The \DOT training split thus includes only the remaining $4,515$ dialogues.

\paragraph{Models} We train two models using T5-3B \cite{2020t5} as a base model: \texttt{T5-D0T} trained on \DOT and \texttt{T5-SGD} trained on SGD. We also compare against the GPT-based automatic annotator used to create silver \DOT state update labels (\texttt{GPT-D0T}). Implementation details presented in Appx. \ref{appx:implemenation_details}.

\vspace{1ex}
\begin{table}[htp!]
    \centering\small
    \resizebox{\columnwidth}{!}{
    \begin{tabular}{@{}lllllll}
        \toprule
        & \multicolumn{3}{c}{\bf \DOT} & \multicolumn{3}{c}{\bf SGD} \\
        \cmidrule(lr){2-4} \cmidrule(lr){5-7}
        \bf Model & \multicolumn{1}{c}{\tt CP} & \multicolumn{1}{c}{\tt CR} & \multicolumn{1}{c}{\tt HM} & \multicolumn{1}{c}{\tt $\:\:$CP} & \multicolumn{1}{c}{\tt CR} & \multicolumn{1}{c}{\tt HM} \\
        \midrule
        \tt T5-SGD & 32.3 & 72.6 & 44.7 & 69.3 & \bf 90.8* & 78.6 \\
        \tt GPT-D0T & 93.3* & \bf 82.0* & 87.3 & 90.0* & 84.7 & 87.3 \\
        \tt T5-D0T & \bf 95.7* & 81.2* & \bf 87.9 & \bf 94.7$\dagger$ & 81.7 & \bf 87.7 \\
        \bottomrule
    \end{tabular}
    }
    \caption{Human evaluation of completeness (\texttt{CP}), correctness (\texttt{CR}), and their harmonic mean (\texttt{HM}) for each dialogue state generator. */$\dagger$ denote statistical significance against unstarred/all results in the same column (Agresti-Caffo, $p < 0.05$).}
    \label{tab:proportions_combined}
    \vspace{-2ex}
\end{table}

\begin{table*}[htbp]
    \centering
    \resizebox{0.9\textwidth}{!}{
    \begin{tabular}{@{}lccccccc|ccccccc}
        \toprule
        & & \multicolumn{6}{c}{\bf MultiWOZ} & & \multicolumn{6}{c}{\bf SGD} \\
         & & \multicolumn{3}{c}{Slot} & \multicolumn{3}{c}{Value} & & \multicolumn{3}{c}{Slot} & \multicolumn{3}{c}{Value} \\
        \cmidrule(lr){3-5} \cmidrule(lr){6-8} \cmidrule(lr){10-12} \cmidrule(lr){13-15}
        \bf Model & C & P & R & F1 & P & R & F1 & C & P & R & F1 & P & R & F1 \\
        \midrule
         \texttt{DSI} & 522 & 96.2 & 80.7 & 87.7 & 41.5 & 57.4 & 37.2 & 11992 & - & - & \bf 92.2 & - & - & 46.2 \\
         \texttt{USI} & 290 & \bf 100.0 & 93.6 & \bf 96.7 & 61.3 & 67.3 & 58.7 & 806 & - & - & 77.0 & - & - & 47.5 \\
         \texttt{GenDSI} & 180 & 85.6 & \bf 96.8 & 90.9 & 81.4 & \bf 70.2 & 70.5 & 746 & \bf 92.4 & 77.9 & 84.5 & 65.4 & \bf 50.0 & \bf 48.8 \\
         \; \footnotesize \texttt{- slot names} & \bf 157 & 73.9 & 90.3 & 81.3 & 85.2 & 47.7 & 55.3 & \bf 467 & 76.4 & 75.6 & 76.0 & \bf 70.6 & 36.3 & 37.9 \\
         \; \footnotesize \texttt{+ all domains} & 161 & 85.1 & \bf 96.8 & 90.6 & \bf 87.9 & 68.1 & \bf 71.0 & 737 & 90.8 & \bf 79.1 & 84.5 & 68.0 & 47.2 & 47.7 \\
        \bottomrule
    \end{tabular}
    }
    \caption{Schema induction results showing Precision/Recall/F1 (P/R/F1) for both induced slots and discovered values, as well as the induced Slot Count (C). Note that the optimal Slot Count would equal the gold slot counts of 31 and 82 for MultiWOZ and SGD respectively. DSI and USI results taken from \citet{yu_unsupervised_2022}.}
    \label{tab:dsi-results}
    
\end{table*}

\paragraph{Results} As shown in Table \ref{tab:proportions_combined}, \texttt{T5-D0T} exhibits the best overall performance, achieving approximately 81\% correct slot-value inferences and completely covering all key information in 95\% of turns. The fact that there was nearly zero performance drop-off on the out-of-distribution SGD evaluation demonstrates its robustness in discovering useful slot-values for new domains. As expected, \texttt{GPT-D0T} exhibits similar performance, as it generated the labels used to train \texttt{T5-D0T}; however, \texttt{GPT-D0T} is much costlier due to multiple API calls to GPT3.5 and GPT4. \texttt{T5-SGD} achieves the highest correctness score when evaluated on held-out SGD domains, but its completeness score of only 70\% demonstrates it is incapable of fully adapting to unseen domains. On the out-of-distribution \DOT evaluation, the performance of \texttt{T5-SGD} heavily suffers, achieving only 32\% completeness and 73\% correctness. This result demonstrates the difficulty of discovering state information in unseen domains, and indicates that SGD is insufficiently diverse as a training resource for this purpose.

\section{Schema Induction Evaluation}
\label{sec:ssi_evaluation}
To evaluate our SSI approach, we use the benchmark defined by \citet{yu_unsupervised_2022} on the validation splits of MultiWOZ 2.1 \cite{eric_multiwoz_2020} and SGD \cite{rastogi_towards_2020} datasets. 
This evaluation method measures the quality of an induced set of slot clusters by matching it against a gold reference slot schema. 

Matching is performed automatically by computing the centroid of each induced and gold reference slot cluster using BERT encodings \cite{devlin-etal-2019-bert} of their values. Each induced cluster is mapped to the gold slot whose cluster centroid is nearest by cosine similarity, or to no cluster if there is no match of 80\% similarity or higher. Similarly, in order to evaluate the purity and coverage of clustered values, discovered values are matched against the gold value labels that fill each slot. This value matching is performed between the values that fill each gold slot and the discovered values of all induced clusters mapped to that slot using fuzzy string matching.

\paragraph{Metrics} Given the mapping of induced clusters to gold slots, \textit{Slot Precision} measures the proportion of induced clusters that were able to be matched to a gold slot, \textit{Slot Recall} is the proportion of gold slots that were matched with at least one induced cluster, and \textit{Slot F1} is their harmonic mean. Since multiple induced slots are allowed to map to a single gold slot, the induced \textit{Slot Count} is also reported to measure redundancy. \textit{Value Precision} is the average proportion of discovered values that matched to gold values, averaged across all gold slots. Similarly, \textit{Value Recall} is the average proportion of gold values that were matched to a discovered value, and \textit{Value F1} is the average F1 score across all gold slots. Equations defining these metrics are presented in Appx. \ref{appx:ssi_metrics}.

\paragraph{Models} Our SSI approach, \texttt{GenDSI}, uses a T5-3B model trained on the \DOT dataset. Since \DOT contains some task domains that are related to domains appearing in MultiWOZ and SGD, we manually review and filter out 34 domains with overlap and train our dialogue state generator on all \DOT dialogues in remaining domains. We also evaluate the performance when using a model trained with all \DOT domains (\texttt{GenDSI +all domains}), which simulates extending our approach using the \DOT data generation method to create synthetic training resources for target domains. Additionally, we evaluate a version of our approach where value candidates are encoded without their predicted slot names (\texttt{GenDSI -slot names}) to measure the benefit of concretely representing value type information. Implementation details provided are in Appx. \ref{appx:implemenation_details}. Finally, we compare to two strong baselines from previous work: 

\begin{itemize}
    \item DSI \cite{min_dialogue_2020}, which leverages a Part-of-Speech (POS) tagger, Named Entity Recognition (NER) tagger, and coreference resolution model to extract value candidate spans using a set of heuristic rules. Slot clusters are then assigned to value candidates using a neural latent variable model.
    \item USI \cite{yu_unsupervised_2022}, which is the SoTA SSI approach. It is a fully unsupervised SSI approach that leverages attention scores between token spans estimated using a pretrained language model to extract value candidates. A three-step hierarchical clustering procedure is then used that aims to cluster value types, then domains, then slots, using HDBSCAN.
\end{itemize}

\paragraph{Results} As shown in Table \ref{tab:dsi-results}, \texttt{GenDSI} outperforms the previous SoTA \texttt{USI} on almost every aspect of the SSI task. It contains fewer redundant slot clusters, superior recall of gold slots, higher cluster purity as measured by value precision, and better coverage of gold slot values. The only metric on which \texttt{GenDSI} did not outperform \texttt{USI} is slot precision on the MultiWOZ evaluation. This is because the state generator model learned to predict boolean slot values from the \DOT dataset that represent intent types, such as greeting and requesting information, which are considered as precision errors under this evaluation since gold slots do not encode intent classes. The performance of \texttt{GenDSI -slot names} dropped considerably on all metrics other than slot count, indicating the utility of inferring concrete slot names when discovering value candidates. Surprisingly, \texttt{GenDSI +all domains} did not afford any meaningful benefit, which may indicate that our approach generalizes to new domains without the need to generate in-domain resources. 

\paragraph{Slot Name Evaluation} Our SSI approach is the first to enable automatic naming of slot clusters. Simply labeling each cluster with the most frequent candidate slot name achieves $93.5\%$ correctly named clusters by human evaluation.

\section{Conclusion}
\label{sec:ssi_conclusion}

This work presents a new SoTA for SSI, demonstrating the power of a generative approach to value candidate discovery. Our dialogue state generator model shows excellent performance for discovering key dialogue state information from unlabeled dialogues without any prior knowledge of the task domain. Its ability to label discovered values with appropriate slot names provides rich type information, allowing a simple clustering method to induce a quality slot schema for unseen domains. Despite this advancement, there is still room to improve SSI. In particular, current SSI methods produce a far greater number of induced slots compared to human-defined schemas. Although our approach reduces the number of induced slots somewhat, future work should aim for SSI with minimal redundancies in induced slots to further improve the utility of SSI in practical settings.


\section*{Acknowledgments}

We gratefully acknowledge the support of the Amazon Alexa AI grant. Any opinions, findings, and conclusions or recommendations expressed in this material are those of the authors and do not necessarily reflect the views of the Alexa AI.


\bibliography{custom}

\appendix
\pagebreak
\newpage

\section{State Generator Evaluation Details}
\label{appx:dsg_metrics}

To facilitate a thorough evaluation of dialogue state generators, a human evaluation measures the following two key aspects:

\paragraph{State Update Completeness} measures the proportion of predicted state updates that humans have judged to fully capture the key information in their associated turns.
Human judges are asked to read each turn within its context and make a binary decision on whether or not \textit{any} essential information is missing in the state update such that:
$$
\texttt{CP} = \frac{1}{|\mathcal{U}|}\sum_{\forall U \in\: \mathcal{U}} \mathbb{I}(\textbf{complete}(U))
$$
$\mathcal{U}$ is a list of all state updates across dialogues to be evaluated and $\mathbb{I}(x)$ is $1$ if $x$ is true; otherwise, $0$.
Note that the judges are not responsible for finding \textit{all} missing information but identifying at least one to assess completeness for efficient evaluation.

\paragraph{Slot Value Correctness} measures the proportion of slot-value pairs that humans have judged to accurately represent \textit{specific} information in their corresponding turns. 
Judges are asked to mark each slot-value pair as correct if it makes sense and is entirely faithful to the content of the associated turn s.t.:
$$
\texttt{CR} = \frac{1}{\sum_{\forall U \in\: \mathcal{U}} |U|}\sum_{\forall U \in\: \mathcal{U}}\sum_{\forall (s,v) \in U} \mathbb{I}(\textbf{correct}(s,v))
$$
Note that both the slot name $s$ and value $v$ must be accurate for $\mathbb{I}(\textbf{correct}(s,v))$ to be $1$.

\section{SSI Evaluation Metrics}
\label{appx:ssi_metrics}

An SSI model produces a list of slot clusters $\hat{S} = [\hat{s_1}, \hat{s_2}, ..., \hat{s_n}]$ where each slot $\hat{s_i}$ is a cluster of values $\hat{s_i} = [\hat{v_1}, \hat{v_2}, ..., \hat{v}_{|\hat{s_i}|}]$. The quality of these slot clusters is measured by matching them against a list of gold reference slots $S = [s_1, s_2, ..., s_m]$, each of which can be similarly represented as a list of the gold value labels that fill each slot such that $s_i = [v_1, v_2, ..., v_{|s_i|}]$.  

Matching is performed by assigning each induced slot $\hat{s_i}$ to one or zero gold slots, creating a mapping $M : \hat{S} \rightarrow S \oplus [\text{none}]$. This matching is performed automatically. First, a centroid representation $c_i$ is computed for each induced slot cluster and each gold slot cluster using the average BERT \cite{devlin-etal-2019-bert} encoding of each value:

$$
    c_i = \frac{\sum_{v_j \in s_i}\textsc{BERT}(v_j)}{|s_i|}
$$

Each induced cluster is mapped to the gold cluster whose centroid is closest by cosine distance, or to \texttt{none} if no gold centroid is within $\geq 0.8$ cosine similarity.

Given the mapping $M$ from predicted to gold slots, the evaluation metrics are calculated follows:

\paragraph{Slot Precision} is the proportion of predicted slots that were able to be mapped to a gold slot:
$$
    \text{SP} = \frac{\sum_{\hat{s_i} \in \hat{S}}{1_S(M(\hat{s_i}))}}{|\hat{S}|}
$$

\paragraph{Slot Recall} is the proportion of gold slots for which there is at least one corresponding predicted slot:
$$
    \text{SR} = \frac{|\{M(\hat{s_i}) : \hat{s_i} \in \hat{S}\} - \{\text{none}\}|}{|S|}
$$

\vspace{10pt}

\paragraph{Slot F1} is calculated normally as the harmonic mean of precision and recall:
$$
\text{S-F1} = 2 \times \frac{\text{precision} \times \text{recall}}{\text{precision} + \text{recall}}
$$

\paragraph{Slot Count} In the above Slot Precision calculation, multiple predicted clusters are allowed to be mapped to a single gold slot. This choice of formulation was made by previous work to avoid punishing the schema induction approach for inducing a finer-grained schema than what the gold schema provides, but fails to reflect the number of redundant clusters that are induced. To mitigate this, the number of induced slots is reported as an additional evaluation metric, where a lower number of induced slots is considered preferable.

\paragraph{Value Precision} is meant to measure the purity of predicted slot clusters. It is calculated only between matched predicted clusters $\hat{S}_{matched}$ and matched gold clusters $S_{matched}$. For each gold slot with at least one match $s_i \in S_{matched}$, the proportion of predicted values in the mapped predicted slots that have a fuzzy match to some gold slot value is measured using fuzzy match boolean function $f$:

$$
\resizebox{\columnwidth}{!}{$
    \text{VP}_{s_i} = \frac{|\{\hat{v}_{kl} : \hat{v}_{kl} \in \hat{v_k}, M(\hat{s_k}) = s_i, v_{ij} \in s_i, f(v_{ij}, \hat{v}_{kl}) \}|}{|\{ \hat{v}_{kl} : \hat{v}_{kl} \in \hat{v_k}, M(\hat{s_k}) = s_i \}|}
$}
$$

\noindent The final Value Precision score is an average across matched gold slots calculated in this way:

$$
    \text{VP} = \frac{\sum_{s_i \in S_{matched}} \text{VP}_{s_i}}{|S_{matched}|}
$$

\paragraph{Value Recall} is calculated similarly to Value Precision. For each gold slot with a mapping to one or more predicted clusters, recall is measured as the proportion of gold values that have a fuzzy match to some value in the corresponding predicted clusters:

$$
\resizebox{\columnwidth}{!}{$
    \text{VR}_{s_i} = \frac{|\{v_{ij} : \hat{v}_{kl} \in \hat{v_k}, M(\hat{s_k}) = s_i, v_{ij} \in s_i, f(v_{ij}, \hat{v}_{kl}) \}|}{|s_i|}
$}
$$

The final Value Recall is also averaged across matched gold slots:

$$
    \text{VR} = \frac{\sum_{s_i \in S_{matched}} \text{VR}_{s_i}}{|S_{matched}|}
$$

\vspace{1ex}
\section{State Generator Evaluation Interface}
\label{appx:generator_evaluation_interface}

\begin{figure*}[htb]
    \centering
    \includegraphics[width=\textwidth]{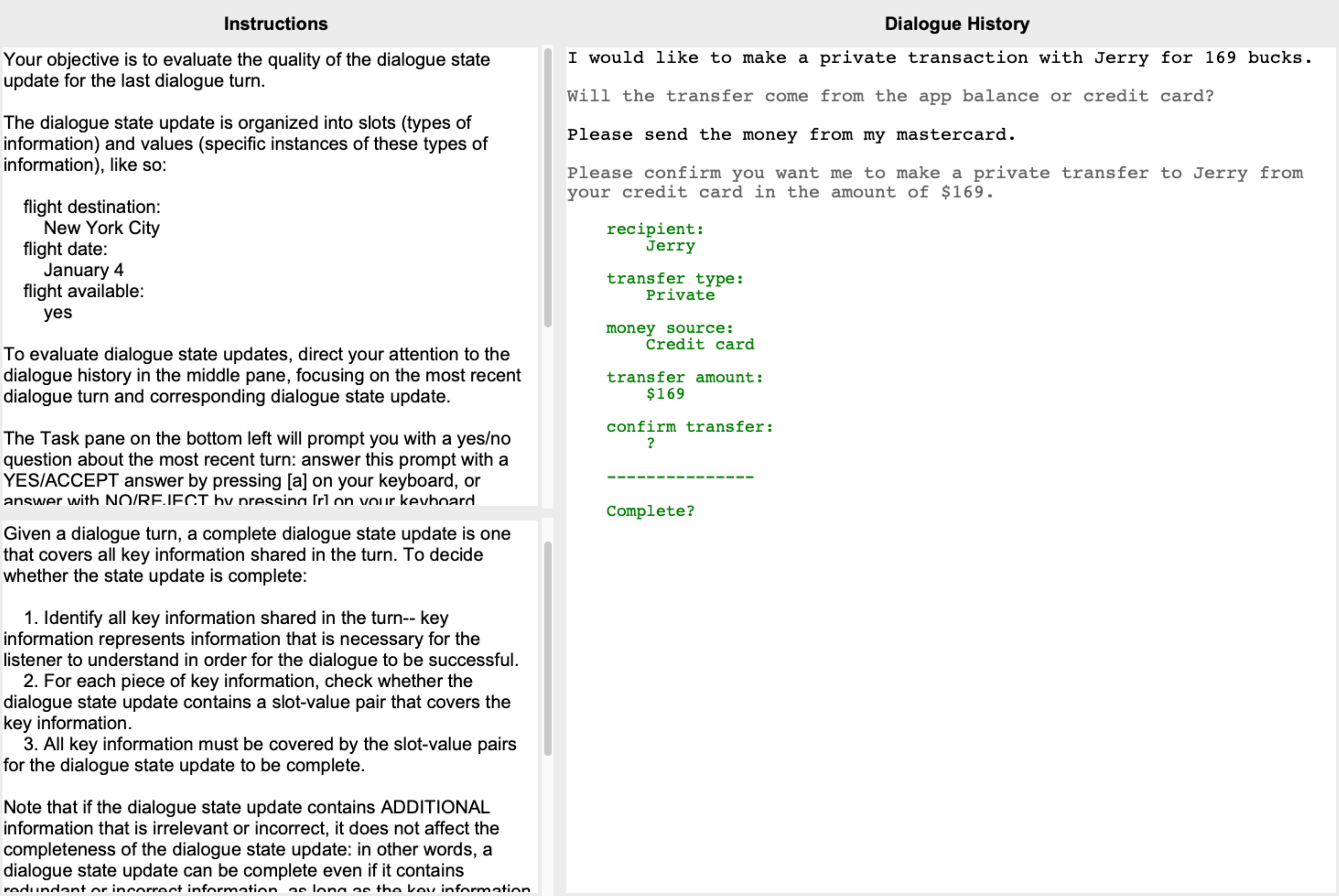}
    \caption{Annotation interface with instructions for human evaluation of Completeness of predicted state updates.}
    \label{fig:completeness-screenshot}
\end{figure*}

\begin{figure*}[htb]
    \centering
    \includegraphics[width=\textwidth]{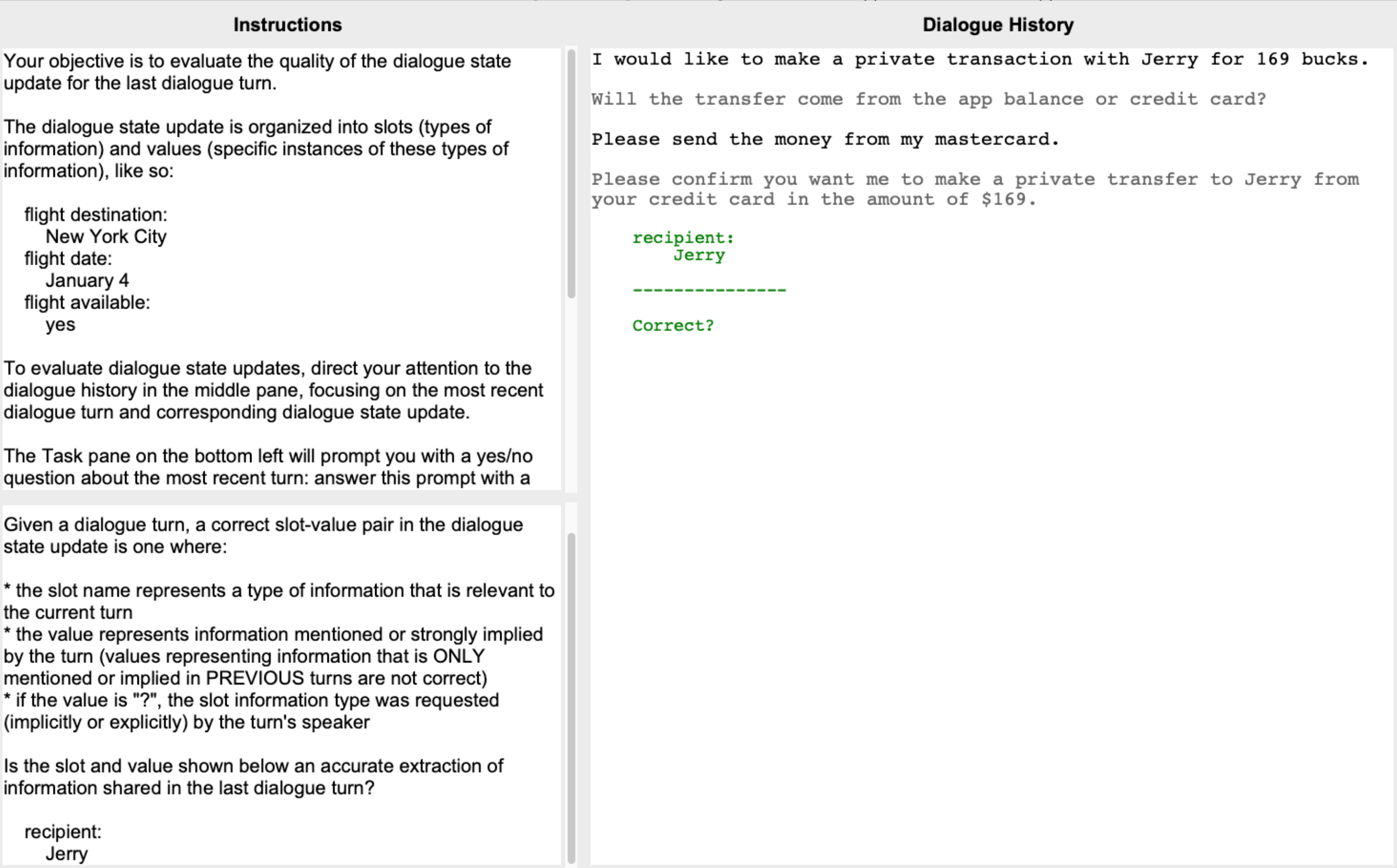}
    \caption{Annotation interface with instructions for human evaluation of Correctness of predicted slot-value pairs.}
    \label{fig:correctness-screenshot}
\end{figure*}

Figure \ref{fig:completeness-screenshot} shows a screenshot of the interface when performing completeness annotations, and Figure \ref{fig:correctness-screenshot} shows a screenshot of the interface when performing correctness annotations. Note that the application interface relies on custom keybindings (e.g. pressing the \texttt{y} or \texttt{n} keys to indicate ``yes" or ``no") for annotators to record their evaluation judgements.

\vspace{1ex}
\section{Implementation Details}
\label{appx:implemenation_details}

\paragraph{Dialogue State Generator} All dialogue state generator models were trained using the original version of T5-3B using the huggingface transformers library\footnote{\url{https://huggingface.co/docs/transformers}}. All training was performed using a learning rate of $1e-4$, weight decay of $5e-3$, batch size $128$, and for exactly $1$ epoch, using the Adam optimizer.

\paragraph{Slot Schema Induction} All SSI models used a T5-3B dialogue state generator model trained with the configuration presented above. The \texttt{all-MiniLM-L6-v2} model from SentenceTransformers\footnote{\url{https://www.sbert.net}} was used for slot-value encoding. All HDBSCAN runs used the CUML\footnote{\url{https://docs.rapids.ai/api/cuml/stable/}} library with a a min.~samples of $5$, minimum cluster size $25$, and cluster merge epsilon $0.3$. 


\end{document}